\documentclass[letterpaper, 10pt, conference]{ieeeconf}
\IEEEoverridecommandlockouts
\usepackage{flushend}
\usepackage{marginnote}
\usepackage{graphics}
\usepackage[pdftex]{graphicx}
\DeclareGraphicsExtensions{.pdf,.png,.jpg}
\usepackage[font={small}]{caption}
\usepackage{subcaption}
\usepackage[rightcaption]{sidecap}
\usepackage{pbox}
\usepackage{mathtools}
\usepackage{amsmath, amssymb, amscd}
\usepackage{ wasysym } %
\usepackage{amsfonts}
\usepackage{mathptmx} %
\usepackage{gensymb} 
\DeclareMathAlphabet{\mathcal}{OMS}{lmsy}{m}{n}
\DeclareSymbolFont{largesymbols}{OMX}{cmex}{m}{n}
\usepackage{textcomp} %
\usepackage[ruled,vlined,linesnumbered]{algorithm2e} 
\usepackage{algorithmicx, algpseudocode} 
\usepackage{array} %
\usepackage{tabularx}
\usepackage{multirow}
\usepackage{multicol}
\usepackage{booktabs}
\usepackage[T1]{fontenc} 
\usepackage[utf8]{inputenc}
\usepackage[english]{babel} %
\usepackage{units}
\usepackage{bm}
\usepackage{times} %
\usepackage{xspace}
\usepackage{flushend}%
\usepackage{csquotes}
\usepackage{makeidx}
\usepackage{soul} %
\usepackage{subfiles} %

\usepackage[protrusion=true,expansion=true]{microtype}
\usepackage[yyyymmdd]{datetime}

\date{\protect\formatdate{1}{1}{2001}}

\usepackage{url}
\makeatletter
\g@addto@macro{\UrlBreaks}{\UrlOrds}
\makeatother
\usepackage{color}
\usepackage[usenames,dvipsnames,table]{xcolor}
\usepackage[pdfborder={0 0 0.5}]{hyperref}
\hypersetup{
    colorlinks=true,
    linkcolor=black,
    citecolor=black,
    filecolor=cyan,
    urlcolor=black
}

\usepackage{amsmath}
\usepackage{balance}
\usepackage{caption}
\numberwithin{equation}{section}

\begin{document}

\title{\LARGE \bf
Inferring the Material Properties of Granular Media for Robotic Tasks
}

\author{Carolyn Matl$^{1, 2}$, Yashraj Narang$^{2}$, Ruzena Bajcsy$^{1}$, Fabio Ramos$^{2, 3}$, Dieter Fox$^{2, 4}$%
\thanks{$^{1}$Department of Electrical Engineering and Computer Science, University of California, Berkeley, CA, USA}%
\thanks{$^{2}$NVIDIA Corporation, Seattle, WA, USA}%
\thanks{$^{3}$School of Computer Science, University of Sydney, Sydney, Australia}
\thanks{$^{4}$Paul G. Allen School of Computer Science \& Engineering, University of Washington, Seattle, WA, USA}
}

\maketitle

\begin{abstract}

Granular media (e.g., cereal grains, plastic resin pellets, and pills) are ubiquitous in robotics-integrated industries, such as agriculture, manufacturing, and pharmaceutical development. This prevalence mandates the accurate and efficient simulation of these materials. This work presents a software and hardware framework that automatically calibrates a fast physics simulator to accurately simulate granular materials by inferring material properties from real-world depth images of granular formations (i.e., piles and rings). Specifically, coefficients of sliding friction, rolling friction, and restitution of grains are estimated from summary statistics of grain formations using likelihood-free Bayesian inference. The calibrated simulator accurately predicts unseen granular formations in both simulation and experiment; furthermore, simulator predictions are shown to generalize to more complex tasks, including using a robot to pour grains into a bowl, as well as to create a desired pattern of piles and rings. Visualizations of the framework and experiments can be viewed at \url{https://youtu.be/OBvV5h2NMKA}. 
\end{abstract}

\IEEEpeerreviewmaketitle

\section{Introduction}

A granular material is a collection of discrete macroscopic particles that primarily experience inelastic collisions and are usually unaffected by temperature \cite{jaeger1992physics}. Naturally, this definition spans a wide variety of materials, from sand and stones to powders and pills to seeds and cereals, making granular material one of the most manipulated substances in the construction, pharmaceutical, agriculture, and food industries \cite{richard2005slow}.
As robots continue to permeate these industries, they will inevitably interact with granular materials in a variety of settings, such as pouring flour in the kitchen, scattering seeds on a farm, or traversing loose terrain \cite{li2013terradynamics}. 

Despite the ubiquity of granular materials, much of the physics behind their complex behaviors remains a mystery.
Granular materials are often considered their own category of matter (separate from solids, liquids, and gases), and consequently, new constitutive laws must be defined. While recent work has delivered promising advances in the formulation of continuum models for granular materials \cite{jop2006constitutive, anand2000granular, dunatunga2015continuum, nemat2002constitutive, kamrin2019non, kamrin2018quantitative}, researchers frequently rely on numerical techniques and simulations to corroborate and extend their predictions.

The state-of-the-art numerical technique for simulating granular materials is the discrete element method (DEM), where each particle is represented independently. Although DEM has primarily been used in physics, it has also been leveraged by the robotics community for applications such as locomotion on sand or gravel \cite{maladen2011undulatory, knuth2012discrete, negrut2013investigating} and motion planning for pouring \cite{dierichs2013robotic}. However, calibrating the material parameters of the simulated grains is still an open problem. Most DEM calibration methods involve a tedious cycle of performing small-scale materials tests, manually tuning parameters, and iterating until the desired experimental behavior is captured. To reduce this burden, the physics community has posed the following challenge: how can particle-scale parameters (e.g., frictional properties) be determined from observations of macroscale behavior (e.g., granular pile shape) \cite{jaegergroup}? Answering this could lead to more efficient methods for refining simulations to match real macroscale behavior. %

\begin{figure}[t]
	\includegraphics[width=\linewidth]{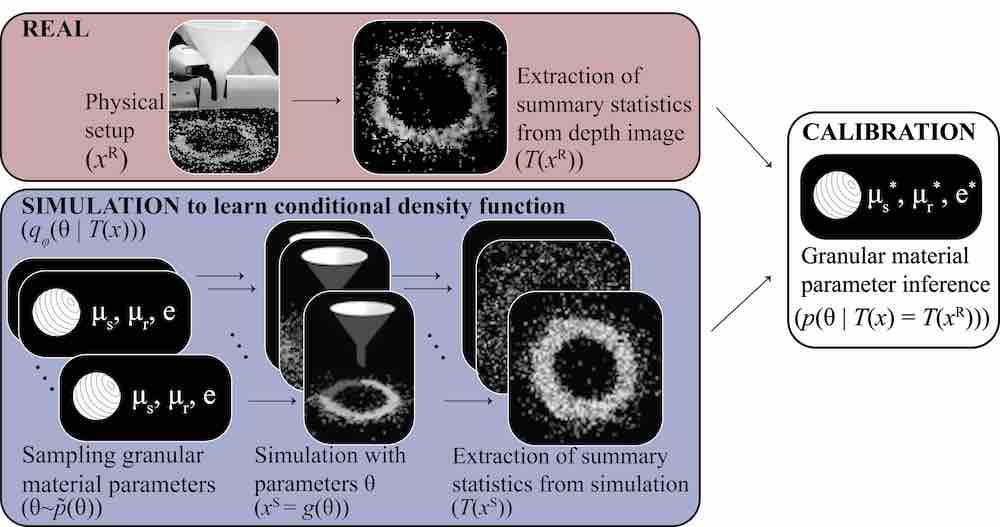}
	\centering
	\caption{Granular parameter inference framework. (Bottom branch): Multiple granular parameter sets are randomly sampled from a prior distribution $\theta \sim \tilde{p(\theta)}$, and for each, summary statistics $T(x^S)$ are extracted from a  simulation $g(\theta)$. The set of $T(x^S)$ are used to learn a conditional density function for the parameters using BayesSim. (Top branch): The summary statistics of a depth image of a real granular formation are extracted. With the conditional density function, a posterior is estimated to infer the likeliest simulation granular parameters to match the macroscopic granular behavior. 
		\label{fig:1}}
\end{figure}

This paper addresses this question by integrating a framework for likelihood-free Bayesian inference (BayesSim \cite{ramos2019bayessim}) with a fast and efficient physics simulation platform capable of DEM simulations (NVIDIA's Isaac Simulator \cite{isaac}), as well as experimental observations using off-the-shelf depth cameras. %
Using this software and hardware framework, depicted in Figure \ref{fig:1}, particle-scale material properties (i.e., coefficients of friction and restitution) of granular materials are inferred from simple, inexpensive observations of macroscale behavior in the real world. The calibrated simulator can then be leveraged to predict unseen granular formations in both simulation and experiment. This paper concludes by demonstrating how this research may be applied to robotics tasks by using the developed framework on an industrial robot to manipulate agricultural grains (i.e., couscous and barley).

The main contributions of this paper are
\begin{itemize}
    \item A software and hardware framework that integrates likelihood-free Bayesian inference, an efficient physics simulator, and depth images of granular formations to estimate particle-scale material properties
    \item A demonstration of the use of this framework to calibrate material properties of agricultural grains (couscous and barley) for a simulator, which is then evaluated against new scenarios in simulation and experiment
    \item A robotic demonstration that uses the proposed framework to reason about agricultural grains in order to complete a simple manipulation task
\end{itemize}

\section{Related Work}

\subsection{Granular Materials in Physics}

The mechanics of granular materials is an active area of research within the physics community. %
Recent advancements present new constitutive relations of granular media that relate flow behavior to particle-scale properties like grain size \cite{kamrin2018quantitative}, geometry \cite{mazhar2014studying}, and surface friction \cite{kamrin2014effect}. Throughout this body of work, DEM simulators are consistently used to evaluate and extend theoretical explanations \cite{cundall1979discrete, zhang2017microscopic, kamrin2014effect, jop2006constitutive, mazhar2014studying}. This emphasis on simulation encourages the development of frameworks that can infer difficult-to-measure material parameters (e.g., frictional coefficients of a grain) from macroscale behavior of collections of grains (e.g., pile shape). This work proposes one such framework, designed primarily for robotic applications due to its speed, efficiency, and minimal hardware requirements.

\subsection{Granular Materials in Robotics}

Within robotics, the mechanics of granular materials has been investigated to design jamming grippers and manipulators \cite{brown2010universal, thompson2015soft, liu2010granular, cianchetti2014soft}, as well as mobile robots that can traverse granular terrain \cite{zhu2019data, li2013terradynamics, maladen2011undulatory, knuth2012discrete, zhang2013ground, negrut2013investigating}. Recent work has explored the perception of granular materials \cite{clarke2018learning, chen2016learning, sinapov2014learning} and their manipulation in scooping \cite{clarke2018learning} and pouring \cite{schenck2017learning} tasks. Such work has largely focused on the classification of granular materials rather than estimation of material parameters; furthermore, it has often been highly task-specific. However, granular parameter identification is important to enable physical reasoning of how different grains may behave in unseen scenarios while dynamically interacting with robots. Due to the complexity of granular physics, researchers in robotics have leveraged simulation to model robot-granular interactions \cite{zhu2019data, zhang2013ground, maladen2011undulatory, negrut2013investigating}. 

\subsection{Simulation of Granular Materials}

 The state-of-the-art numerical technique to simulate granular materials is the Discrete Element Method (DEM), where each grain is generally represented as an independent particle that interacts with others only through direct contact. Other granular simulation techniques include the Material Point Method (MPM) \cite{hu2018moving, dunatunga2015continuum}, which constructs a background mesh to compute forces, and Position Based Dynamics (PBD) \cite{muller2007position, macklin2013position, macklin2014unified}, which directly manipulates particle positions to enforce attachments and resolve collisions, typically trading off accuracy for speed. Although accurate, DEM often uses explicit time-integration, which can require impractically small time steps for stiff systems. In this work we model contact as a nonlinear complementarity problem (NCP) which is solved implicitly using a GPU-optimized non-smooth Newton solver in the NVIDIA Isaac Simulator platform \cite{isaac, macklin2019non}. This enables both high accuracy and rapid simulation of $\sim$2000 spherical grains. Furthermore, the platform allows direct input of particle-scale parameters such as geometry and material properties, as well as a programmatic interface in which new testing scenarios can be rapidly created. The platform has recently been applied to reinforcement learning applications for non-DEM simulations \cite{chebotar2019closing, liang2018gpu}. %

\subsection{Automatic Calibration of Granular Material Simulation}

Most techniques used to calibrate the material parameters for DEM simulation require tedious experiments and manual tuning until the desired experimental behavior is captured \cite{coetzee2017calibration}. Thus, recent work aims to automate this process. For instance, in \cite{do2017automated}, a genetic algorithm and direct optimization are used to determine the frictional coefficients of sand grains by measuring discharge time from an hourglass. In \cite{benvenuti2016identification}, an artificial neural network is used to infer multiple material parameters of iron ore sinter; the network greatly increases the speed of inference at the expense of training time. Recently, likelihood-free inference techniques, such as Approximate Bayesian Computation (ABC) methods, have been explored to increase the interpretability of results, as well as to enhance robustness to non-unique mappings between material properties and macroscale behavior. Specifically, in \cite{cheng2019iterative}, an iterative Bayesian filtering framework was used to infer the coefficients of friction of glass beads based on 100 measurements of porosity and stress per iteration. %

In the present work, we leverage a likelihood-free Bayesian inference framework called BayesSim \cite{ramos2019bayessim}. The framework allows the complexities of the simulator to be fully abstracted from the inference process and requires only a single training phase followed by a prediction step. This is in contrast to multiple prediction steps necessary in typical Bayesian filtering. Because BayesSim is an $\epsilon$-free method~\cite{papamakarios2016fast} (i.e., it does not rely on sampling with acceptance region defined by $\epsilon$), it is generally more sample-efficient and produces more accurate parametric approximations of the posteriors than ABC methods \cite{papamakarios2016fast}. Furthermore, the preceding studies measured macroscopic behavior using highly-specific experimental setups and advanced laboratory equipment, which may be prohibitively expensive for many robotics applications. In contrast, the present work measures macroscopic behavior (e.g., granular formations from pouring grains through a funnel) using single images taken with inexpensive, off-the-shelf depth cameras.

\section{Problem Statement}
\label{Problem Statement}

\subsection{Simplifying Assumptions}
\label{Assumptions}
This section defines the key simplifications made in the simulator. %
 First, we consider grains with a length scale of at least 1 mm, thus excluding powders. Given that the average size and mass of a grain can be measured, they are assumed as known inputs to the simulator.  While natural grains will represent a distribution of sizes and masses, we make the simplification that all grains are of the same size and mass.

\begin{figure}[t]
	\includegraphics[width=\linewidth]{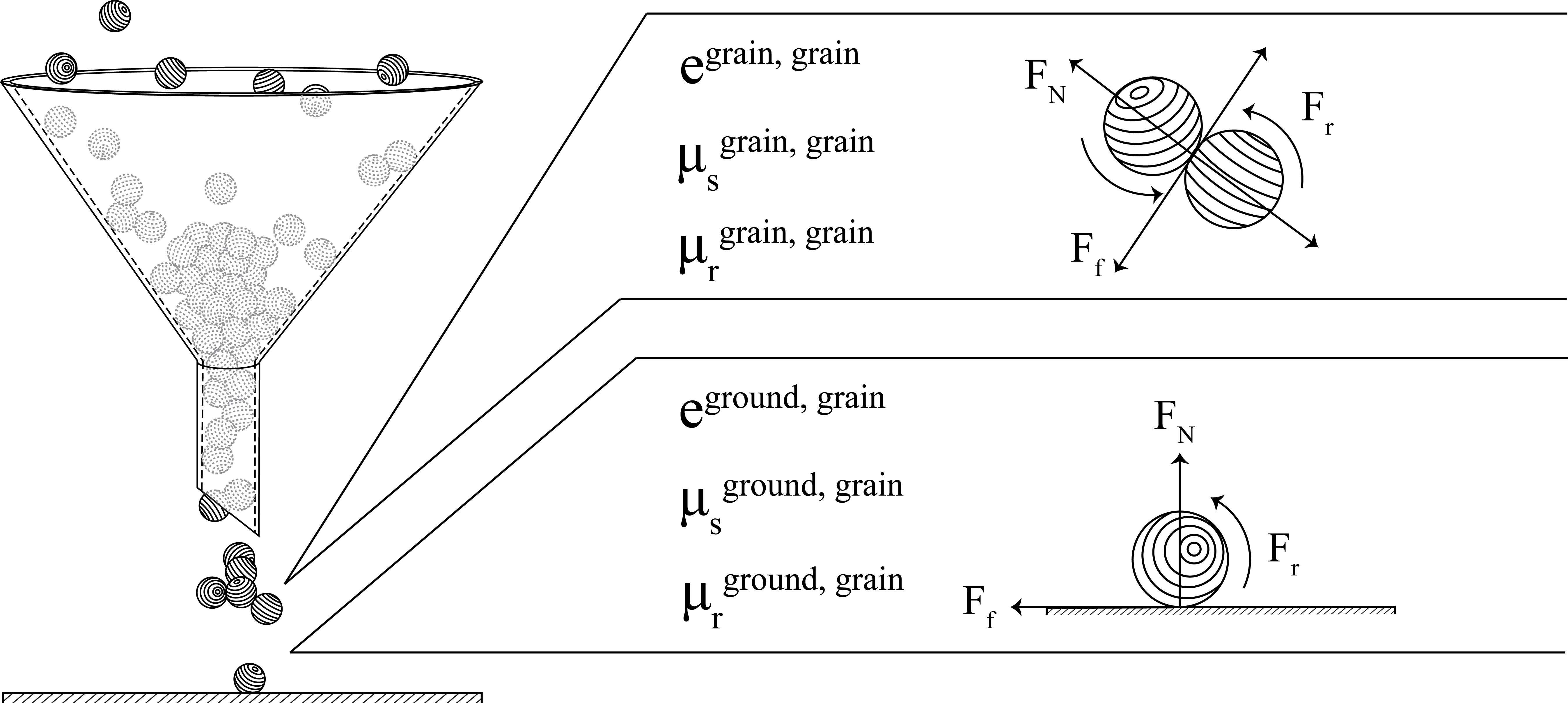}
	\centering
	\caption{The simulations capture sliding friction forces ($F_f$), rolling friction forces ($F_r$), and energy loss due to inelastic collisions for grain-grain and grain-ground interactions. These forces are parameterized by the coefficients of sliding friction ($\mu_s$), rolling friction ($\mu_r$), and restitution ($e$), respectively.
		\label{fig:physics}}
\end{figure}

We further assume that the grains are perfectly rigid, cohesionless, and experience negligible drag, so that the only sources of energy dissipation are friction and collisions with each other and the ground. The forces experienced during these interactions, depicted in Figure \ref{fig:physics}, are functions of the coefficients of restitution $e$, sliding friction $\mu_s$, and rolling friction $\mu_r$ for the grain-to-grain or grain-to-ground interactions. To simplify parameter search, we assume that the grain-to-grain and grain-to-ground coefficients are equivalent. Together, these three parameters enable the simulator to generate a diversity of realistic granular behavior. Using a subset of $\{e, \mu_s, \mu_r\}$ restricts the simulator's expressiveness and may likely cause the real and simulated distributions of granular formations to be completely distinct. 

Finally, we represent all grains as spheres. While it has been shown that aggregate behavior is dependent on grain geometry and size, this coarse approximation makes collisions easier and faster to compute, and thus practical for robotic applications. Further, although the parameters are bounded by real-world values,
it is not the goal of this paper to estimate physically-exact friction and restitution coefficients. Rather, the calibration framework aims to accurately simulate macroscopic behavior of the granular material. For non-spherical grains, larger or smaller valued material parameters may sufficiently compensate for a spherical approximation.

\subsection{Definitions}
\label{Definitions}

Let $\theta = [ \mu_s, \mu_r, e]^T$ represent a vector of the grain simulation parameters, composed of the coefficients of sliding friction $\mu_s$, rolling friction $\mu_r$, and restitution $e$. %
Let $X$ represent the birds-eye-view depth image of the grains once they have reached steady-state. Variables $X^r$ and $X^s$ are used to distinguish real from simulated depth images, respectively, and $X^s = g(\theta)$, where $g(\theta)$ is the simulation. %
Because of the high-dimensional space of depth images $X^r$ and $X^s$, we define the random variable $T(X)$ as a vector composed of statistics (such as average distance of grains from the centroid) to summarize the observed state $X$. 
Let the likelihood function be defined as $p(T(X)|\theta)$ and the posterior be defined as $p(\theta|T(X)=T(X^r))$. The likelihood function, defined implicitly by $g(\theta)$, is often intractable to evaluate or unavailable with such complex simulations, thus motivating the use of a likelihood-free inference approach. 

\subsection{Objective}
\label{Objective}

Given samples of $T(X^s)$ and the corresponding parameters used to generate the simulations, we would like to approximate the posterior function $p(\theta|T(X)=T(X^r))$ such that, from a real observation $X^r$, we can infer granular simulation parameters $\theta$ from the macroscopic behavior of the material, summarized by features $T(X)$. Because DEM simulations leverage various numerical solvers and are associated with complex systems of differential equations, we choose a simulator-agnostic inference framework that approximates the intractable posterior by sampling from the simulator. 
The likelihood-free Bayesian inference framework BayesSim \cite{ramos2019bayessim}, discussed in the following section, is used to approximate this posterior.

\section{Methods}
The goal of this paper's proposed framework is to estimate granular parameters $\theta = [ \mu_s, \mu_r, e]^T$ from summary statistics $T(X^r)$ using BayesSim, which approximates the posterior $p(\theta|T(X)=T(X^r))$ by sampling from a simulator $g(\theta)$. In this section, we discuss the BayesSim likelihood-free framework, simulator $g(\theta)=X^s$, physical experimental setup to observe $X^r$, and extraction of summary statistics $T(X)$.

\subsection{Granular Parameter Inference Using BayesSim}

BayesSim directly learns a conditional density function, $q_{\phi}(\theta|T(X))$ (where $\phi$ is the set of learnable parameters of $q$), by mapping summary statistics to a mixture of $K$ Gaussian components via random Fourier features \cite{rahimi2008random}. The Mixture Density Random Fourier Feature (MDRFF) network learns from $N$ pairs of parameters and summary statistics, $\{\theta_i, T(X^s_i)\}^N_{i=1}$, 
generated by independently sampling from a predefined prior distribution, $p(\theta)$, and forward-simulating each sample, $\theta_i$. For twice differentiability of $q$, the Mat\'{e}rn kernel is used with $\nu=2.5$, where $\nu$ increases with smoothness. The posterior is then parametrically approximated using the learned conditional density $q_\phi(\theta|T(X))$ and a real observation $T(X^r)$. The main advantages of using BayesSim over classical optimization methods are that the complexities and inaccuracies of the DEM simulator and PDE solver are incorporated in the estimated posterior. The multimodality and uncertainty of the approximated posterior can be used to analyze the accuracy with respect to the parameters $\theta$ and robustness with respect to measurement noise. In addition, because BayesSim is an $\epsilon$-free method, it improves on ABC methods by being more sample efficient and producing more accurate posterior approximations \cite{papamakarios2016fast} with a limited number of samples. After the initial cost of running $N$ simulations and training the model, BayesSim can quickly produce posterior approximations, making it a compelling method for granular material inference problems in robotic applications.  

\subsection{Simulation}

Critical dimensions are made equivalent in the simulated and physical scenes. The simulated funnel, which was modeled in 3D based on the physical funnel, is positioned at an equivalent distance above the ground surface.  The spherical grains are scaled to match the diameter of a single grain of Israeli couscous along its longest axis (4 mm). For simplicity, at the start of the simulation, the 2000 grains are each spaced at 8 mm intervals in a 10 $\times$ 10 $\times$ 20 grid, where the bottom grains are just below the top of the funnel. %

The coefficient of restitution $e$ is the ratio of a grain's speed before and after a collision; this coefficient ranges from 0 to 1, corresponding to perfectly inelastic and elastic collisions, respectively. The ranges of the coefficients of sliding friction $\mu_s$ and rolling friction $\mu_r$ that create a noticeable diversity in granular formations are 0 to 1 and $1E\text{-}7$ to $1E\text{-}1$, respectively. For this reason, rolling friction is sampled from a logarithmic distribution. A simulated depth camera is positioned at a birds-eye viewpoint, 29 cm from the surface ground, giving it a 58 cm $\times$ 58 cm field of view. 

The solver chosen is NVIDIA Isaac's Preconditioned Conjugate Residual (PCR) solver, and 20 inner iterations with 4 outer iterations are used in experiments \cite{macklin2019non}. The simulator runs with time step $\nicefrac{1}{60}$ sec, 10 substeps, and a relaxation factor of 0.75 for stability and convergence.
    
\subsection{Physical Setup}

For the physical experiments, the funnel is positioned with its bottom tip 12 cm above the ground surface. The surface is covered in thin, moderate-friction black velvet to limit the spread of the grains upon impact, as well as to simplify segmentation. The primary grain used in the experiments is Israeli couscous, which has the approximate geometry of a rounded cylinder; barley, which has a similar density but is slightly larger and more ellipsoidal, is also used in the latter half of the experiments.
An Intel RealSense D435 depth camera is positioned at 40 cm above the ground surface. This position was chosen to balance the trade-off between field-of-view and clarity of the depth image. 
Although the raw depth images are still noisy at 40 cm, eliminating the background using color segmentation results in well-defined point clouds of the granular formations.  An ABB YuMi dual-arm robot is used for the robot experiments.
    
\subsection{Summary Statistic Extraction}
\label{sec:summarystatistics}
A majority of DEM calibration work relies on extracting specific features (e.g., angle of repose) of a conical pile in order to infer a single parameter (e.g., $\mu_s$). However, initial experiments showed that pouring grains can result in piles or rings, which may be more informative of other material parameters. Moreover, for shallow formations, it may be difficult to measure angle of repose from a depth image. We take a more generalizable approach and propose using a vector of 16 summary statistics to serve as $T(X)$.

To extract the summary statistics features from a granular formation, the depth image $X$ is first downsampled by factor $n$. For $X^s$, $n=20$ and for $X^r$, $n=10$ to match resolutions. Because the $X^r$ are fairly noisy, a color threshold mask is used to segment pixels corresponding to the granular formations. These pixels are then reprojected into 3D using the camera intrinsics matrix. For the real depth image, a plane is fit to reprojected pixels corresponding to the ground surface. The orientation of the fitted ground plane informs the centering and re-orientation of the granular point cloud to account for camera tilt. Let $r$ represent the vector of radial distances of grain points in the horizontal plane from the centroid of the granular formation and $z$ be the vector of grain point heights.

The summary statistics of a single depth image $T(X)$ are chosen as $[max(z), \mu(z), \sigma(z), max(r), \mu(r), \sigma(x), \sigma(y), \sigma(r),$ $IQR(x), IQR(y), IQR(r), KURT(r), dCor(r, z), df, b, A]^T$, where $\mu(\cdot)$, $\sigma(\cdot)$, $IQR(\cdot)$, $KURT(\cdot)$, and $dCor(\cdot, \cdot)$ are the mean, standard deviation, interquartile range, kurtosis, and distance correlation. A chi-distribution, chosen for its likeness in shape to a cross-section of a granular formation, is fit to $r$. Statistics $df$, $b$, and $A$ denote degrees of freedom, shift, and scale, respectively, from a standardized chi probability density. These statistics were chosen to summarize the dispersion and statistical dependence of the grain locations. A preliminary ablation study suggested that subsets of $T(X)$ were sufficiently descriptive in capturing the granular distributions, unless the subsets contained less than three features or the features exclusively pertained to either $r$ or $z$.

\section{Sensitivity Tests for System Design}
\label{sec:sensitivity}

\begin{figure}[t]
	\includegraphics[width=\linewidth]{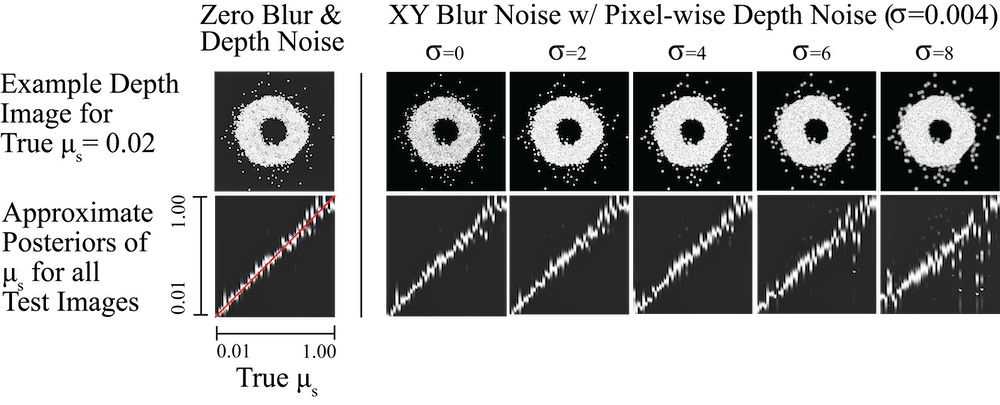}
	\centering
	\caption{Sensitivity to observation error. An example $X^s$ is shown perturbed by different levels of noise. 50 different test images $X^s$ were generated by evenly sampling $\mu_s \in [0.01, 1.0]$. The resulting approximated posteriors are plotted per true value of $\mu_s$, where the highest peaks correspond to the brightest points. A perfect inference would result in a white line overlapping the red. While the model is fairly robust to the level of per-pixel Gaussian depth noise applied, accuracy is more affected by Gaussian blur. Posterior peaks become wider, and distributions can become multimodal with $\sigma>4$. 
		\label{fig:observationerror}}
\end{figure}

\subsection{Preliminary Tests for System Noise}

When running the simulator to train BayesSim, there are two main sources of uncertainty. The first is purely algorithmic, due to the non-determinism of the simulator and numerical error. To measure algorithmic uncertainty, the simulator was run 1000 times with identical conditions except for sliding friction, where $\mu_s \in \{0.01, 1\}$, evenly split. The measured summary statistics for each $\mu_s$ had variances of at least two orders of magnitude smaller than their mean values. 

Second, an insufficient number of samples for training and interpolation uncertainty may contribute to poor performance. To test these, a training set was created by sampling $\mu_s$ uniformly between 0.01 and 1. A 50-simulation testing set was generated by sampling between 0.01 and 1, independently of the training set. Supporting intuition, the accuracy of the friction coefficients inferred by BayesSim improved with more samples. For a training set of 500 samples, the error in the friction coefficient estimates is $0.0191\pm0.0799$, while for 1000 samples, the error is reduced to $0.0062\pm0.0018$.

\subsection{Sensitivity to Observation Error}

 A key assumption of our system is that the summary statistics of the real and simulated depth images will lie within the same distribution. However, the two might not align or overlap given drastically different observations (e.g., due to a low-quality experimental depth image). To test the effect of observation noise on inference, we simulate two types of camera noise on a 50-sample testing set, where $\mu_s$ is evenly sampled from 0.01 to 1 with all other parameters equivalent. An XY Gaussian blur with $\sigma=\{0, 2, 4, 6, 8\}$ is convolved with the depth image, and pixel-wise Gaussian noise with $\sigma=\{0, 0.001, 0.002, 0.003, 0.004\}$ is added to each depth value, with ranges chosen to mimic realistic noise magnitudes.
These two perturbations are applied independently, and the resulting posteriors per true friction value are illustrated in Figure \ref{fig:observationerror}. As shown, inference of $\mu_s$ is fairly robust to pixel-wise noise, but is sensitive to high values of Gaussian blur.

\subsection{Propagation of Error}
\label{sec:propagation}
Given an observation $X$, the approximated posterior encodes the uncertainty of the inferred parameters $\theta^*$. Sensitivity tests were performed to investigate how this uncertainty propagates to errors in the summary statistics $T(g(\theta^*))$. Fixing all other parameters, $\mu_s$, $\ln(\mu_r)$, and $e$ required errors of $\pm 0.05\text{, }3\text{, } 0.01$, respectively, to perturb the height and average radial distance to more than the width of a single grain. For real-world context, manufacturing applications may require that the diameter of a granular pile be predicted to within 5\% of the true value. (For example, for a pile with base diameter equal to 50 times a grain diameter of 4 mm, an error of 1 cm may be allowable). With the proposed system, this specification is fulfilled by the optimal values, with allowable variation of $\pm 0.1\text{, }4 \text{, }0.05$ that is within the performance achieved in the following experiments.

\section{Results}
\label{sec:results}
\subsection{Sim-to-sim Parameter Estimation}
\begin{figure}[t]
	\includegraphics[width=\linewidth]{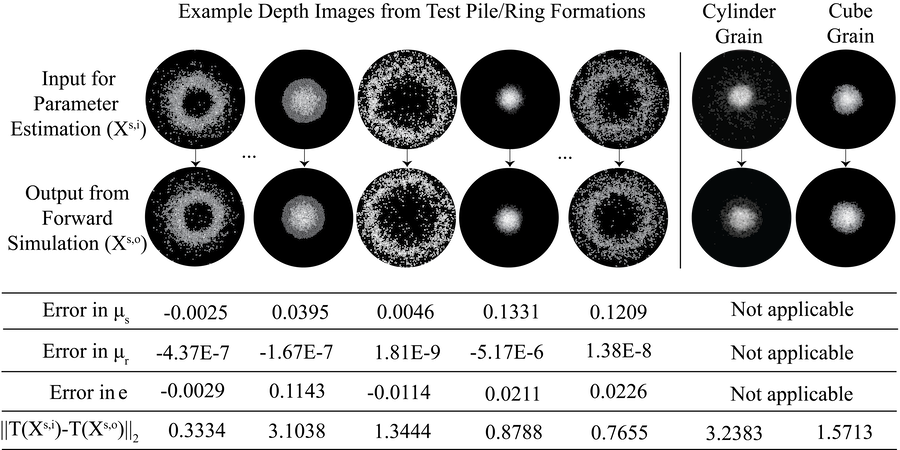}
	\centering
	\caption{Examples comparing simulated depth images of testing set and simulated depth images from running a forward simulation with inferred parameters. Table shows corresponding errors in inferred parameters, as well as L2 errors between summary statistics.
		\label{fig:simtosim}}
\end{figure}

The first objective was to test the accuracy of the trained BayesSim model on simulated depth images. Given a simulated input image $X^{s,i}=g(\theta)$, we aimed to quantify how close the inferred parameters $\theta^* = {\mu_s^*, \mu_r^*, e^*}$ were to $\theta$. 
Furthermore, to evaluate the propagation of error in $\theta^*$, a forward simulation was run to obtain a new pile $X^{s,o}=g(\theta^*)$. The difference between the piles was evaluated via the L2-norm error between $T(X^{s,i})$ and $T(X^{s,o})$, both standardized over the summary statistics of the training dataset. Note that, for some applications, evaluating particular subsets of the summary statistics might be more useful (e.g., height of the granular pile or the angle of repose). To enable generality, all 16 summary statistics defined in Section \ref{sec:summarystatistics} are used to calculate the L2-norm error. We avoid evaluating accuracy via direct comparisons of the downsampled depth images, as the computed error would be highly sensitive to misalignments of the scattered grains rather than reflect meaningful differences in spatial distributions.

The 50 test grain formations were simulated by randomly sampling $\mu_s$ and $e$ uniformly and $\mu_r$ logarithmically from their given ranges. The average error of the inferred $e$ was $-0.0008\pm0.0148$, while that of $\mu_s$ and $ln(\mu_r)$ were $0.0380\pm0.0646$ and $-1.1695\pm0.7032$, respectively. As in Section \ref{sec:propagation}, these error bounds correspond to negligible changes in the resulting statistics. As shown in Figure \ref{fig:simtosim}, forward simulations of the inferred parameters for the 50 test formations generate visually equivalent shapes. To consider as a baseline for the following experiments, the average L2-norm error between the 50 pairs of $\{T(X^{s,i}),  T(X^{s,o})\}$ was $1.582\pm0.3356$. %

Sim-to-sim comparisons also enable a principled approach to testing sensitivity to model error. For instance, the grains are spherical, when in reality most grains are not. We tested sensitivity to mismatched geometry by creating a testing set of cylindrical and cube-shaped grains, where the diameter and length of each grain were 4 mm, the diameter of the training set spheres.
Each testing set consisted of 50 different formations which varied in $\mu_s$, while $e$ and $\mu_r$ were fixed at 0.5 and $1E{-6}$, respectively. Parameters $\mu_s$, $\mu_r$, and $e$ were inferred from the test formations and forward simulated with spherical grains. Comparing $T(X^{s,i})$ (with cylindrical and cubic grains) with $T(X^{s,o})$ (using spherical grains) the L2-norm errors were $4.0139\pm1.0574$ and $1.8301\pm1.9275$, respectively. Spherical grains can approximate formations made with cubes better than cylinders, possibly due to the asymmetry of cylinder geometry. %
Depending on the application, a spherical representation may be sufficient.

\subsection{Real-to-sim Parameter Estimation}

The goal of this experiment was to test whether material parameters of real-world granular formations could be inferred to produce accurate simulations of macroscopic behavior. The initial test set was composed of real depth images of $\sim$2000 grains of Israeli couscous, poured through a funnel fixed at 12 cm above the ground surface. Ten different pours were executed, and for each, ten consecutive depth images were collected and averaged to mitigate temporal noise from the camera. For each of the averaged images $\{X^{r,i}\}^{10}_{i=1}$, summary statistics $\{T(X^{r,i})\}^{10}_{i=1}$ were computed, and $\mu_s$, $\mu_r$, and $e$ were inferred. Forward simulation was executed for each parameter set, and the L2-error was computed between the summary statistics of the real and forward-simulated granular formations. From the ten parameter sets inferred, the five corresponding to the lowest errors were averaged. The inferred parameters for couscous were  $\mu_s=0.6687$, $\mu_r=8.1506E\text{-}7$, and $e=0.7689$, and the parameters for barley were $\mu_s=0.3807$, $\mu_r=1.0613E\text{-}6$, and $e=0.4792$.

\begin{figure}[t]
	\includegraphics[width=\linewidth]{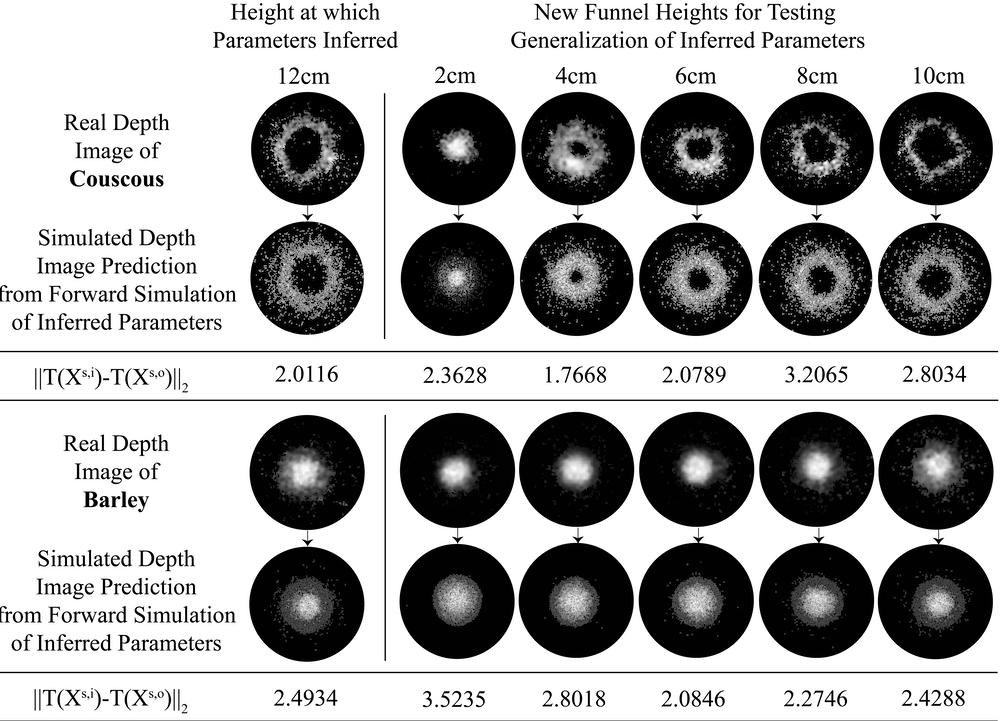}
	\centering
	\caption{Comparisons of real depth images of piles poured at different heights and simulated depth images generated from running a forward simulation at the corresponding heights, using parameters inferred from a height of 12 cm. L2 errors are listed below the depth images.
		\label{fig:realtosim}}
\end{figure}

Forward simulation with the estimated parameters captures similar macroscopic behavior as the five real-world pours at 12 cm, with an average L2-norm error of $2.2263\pm0.1377$ and $2.0311\pm0.3807$ for the couscous and barley formations, respectively. However, it is critical to test how well these parameters generalize to pours at different heights. Five more pours of each grain were performed, with each pour at a different height: $\{$2, 4, 6, 8, 10$\}$ cm off the ground. For each of these pours, a forward simulation was run with the same grain parameters while shifting the funnel height. Figure \ref{fig:realtosim} illustrates that the inferred parameters generalize reasonably well, with relatively low L2 errors. This suggests that it may be possible under this framework to reason about pour heights to create desired pile and ring shapes given a specific type of grain. We test this hypothesis in the following experiment.

\subsection{Robotic Demonstrations}
\begin{figure}[t]
	\includegraphics[width=\linewidth]{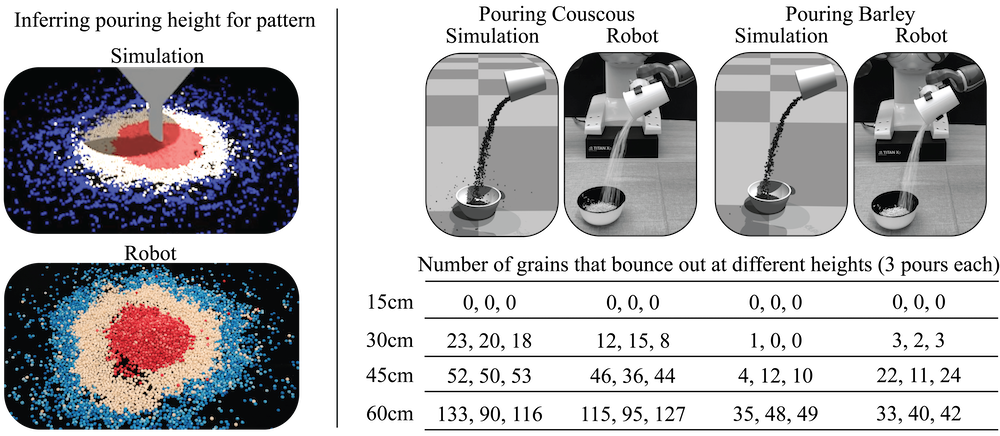}
	\centering
	\caption{Two granular manipulation tasks were tested. (Left): The height of the funnel is inferred to create a desired ring shape. Performance is demonstrated by creating a pattern of three concentric rings. (Right): The calibrated simulator is used to recreate a real-life scene of pouring grains into a bowl. At various heights, both the simulations of couscous and barley reasonably estimate the number of grains that leave the cereal bowl. 
		\label{fig:robotdemos}}
\end{figure}

The first demonstration evaluated whether the presented framework could enable an industrial robot to pour granular material into a desired shape, as may be necessary in a kitchen or factory. A new BayesSim model was trained on 1000 simulated piles. The simulator parameters $\mu_s$, $\mu_r$, and $e$ were set to their inferred values for couscous ($0.6687$, $8.1506E\text{-}7$, and $0.7689$, respectively), and the funnel height was uniformly sampled between 1 and 13 cm. Two summary statistics were computed from the resulting granular formation (i.e., the 5th and 50th percentile of granular radial distance, which approximated the inner and outer radii of a ring). BayesSim was then evaluated on a testing set to infer the funnel height from these two summary statistics; the height was estimated within $0.0530\pm0.2221$ cm. For the demonstration, summary statistics were chosen to correspond to a desired pattern of two concentric rings with an inner pile. BayesSim inferred corresponding funnel heights of $\{$27.1, 10.1, 1.5$\}$ cm. The robot was commanded to pour couscous from these heights, resulting in the desired pattern (Figure \ref{fig:robotdemos}).

The second demonstration evaluated how closely the presented framework could predict undesirable spilling during a granular pouring task. The calibrated simulator was used to simulate the pouring of couscous and barley into a cereal bowl, with the simulation relaxation constant increased from $0.75$ to $0.9$ to ensure accurate simulation of high-speed collisions. The number of grains that spilled out of the bowl was counted. The industrial robot was then used to precisely repeat the experiment in the real-world, pouring grains into a velvet-lined cereal bowl to match simulator conditions. As tabulated in Figure \ref{fig:robotdemos}, in most cases, the calibrated simulator predicted the number of spilled grains across different pouring heights with surprising accuracy. (Please see the supplementary video.)

\section{Conclusions}

In this paper, the material parameters of granular materials were inferred using a new framework combining likelihood-free Bayesian inference, efficient simulation, and simple experiments. The use of GPU-based simulation and off-the-shelf depth cameras may be particularly appealing for robotics applications. Simulation-to-simulation inference was highly accurate, and simulation-to-experiment inference trailed closely in performance. Robotics demos showed that the inferred parameters generalized well to different pouring heights, and furthermore, that a robot can effectively reason about granular material to pour desired granular formations and predict its behavior in dynamic scenarios.

There are numerous exciting opportunities for future work in perceiving and reasoning about granular materials. To improve inference, observations can be amended to include temporal sequences of depth images, with additional summary statistics to capture transient dynamics. To enhance simulation, NVIDIA's Isaac Simulator can be applied to model cohesion, allowing liquid-solid interactions (e.g., capillary bridging); in addition, simulated particle geometries could be represented as superquadrics \cite{podlozhnyuk2017efficient}, enabling more accurate approximation of complex real-world grain geometries. Finally, more elaborate robotic manipulation tasks can be explored, such as moving the end effector along a trajectory to create an asymmetric grain trail, or scooping and pouring granular material into and out of assorted containers.

{\footnotesize
\section*{Acknowledgment} The authors were supported in part by the National Science Foundation Graduate Research Fellowship. We thank Miles Macklin, Renato Gasoto, and Hammad Mazhar for their tireless assistance with the NVIDIA Isaac Simulator, as well as Matthew Matl, Karl Van Wyk, and Qian Wan for their insightful feedback and suggestions.
}


\begin{thebibliography}{10}

\bibitem{jaegergroup}
Granular material by design.
\newblock
  \url{https://jfi.uchicago.edu/~jaeger/group/Granular_Matter_by_Design/Granular_Matter_by_Design.html}.
\newblock Accessed:2019-09-15.

\bibitem{isaac}
{NVIDIA} {I}saac: Virtual simulator for robots.
\newblock
  \url{https://www.nvidia.com/en-us/deep-learning-ai/industries/robotics/}.
\newblock Accessed: 2019-09-11.

\bibitem{anand2000granular}
L~Anand and C~Gu.
\newblock Granular materials: constitutive equations and strain localization.
\newblock {\em Journal of the Mechanics and Physics of Solids},
  48(8):1701--1733, 2000.

\bibitem{benvenuti2016identification}
L~Benvenuti, C~Kloss, and S~Pirker.
\newblock Identification of {DEM} simulation parameters by artificial neural
  networks and bulk experiments.
\newblock {\em Powder technology}, 291:456--465, 2016.

\bibitem{brown2010universal}
Eric Brown, Nicholas Rodenberg, John Amend, Annan Mozeika, Erik Steltz,
  Mitchell~R Zakin, Hod Lipson, and Heinrich~M Jaeger.
\newblock Universal robotic gripper based on the jamming of granular material.
\newblock {\em Proceedings of the National Academy of Sciences},
  107(44):18809--18814, 2010.

\bibitem{chebotar2019closing}
Yevgen Chebotar, Ankur Handa, Viktor Makoviychuk, Miles Macklin, Jan Issac,
  Nathan Ratliff, and Dieter Fox.
\newblock Closing the sim-to-real loop: Adapting simulation randomization with
  real world experience.
\newblock In {\em 2019 International Conference on Robotics and Automation
  (ICRA)}, pages 8973--8979. IEEE, 2019.

\bibitem{chen2016learning}
Carolyn~L Chen, Jeffrey~O Snyder, and Peter~J Ramadge.
\newblock Learning to identify container contents through tactile vibration
  signatures.
\newblock In {\em 2016 IEEE International Conference on Simulation, Modeling,
  and Programming for Autonomous Robots (SIMPAR)}, pages 43--48. IEEE, 2016.

\bibitem{cheng2019iterative}
Hongyang Cheng, Takayuki Shuku, Klaus Thoeni, Pamela Tempone, Stefan Luding,
  and Vanessa Magnanimo.
\newblock An iterative {B}ayesian filtering framework for fast and automated
  calibration of {DEM} models.
\newblock {\em Computer Methods in Applied Mechanics and Engineering},
  350:268--294, 2019.

\bibitem{cianchetti2014soft}
Matteo Cianchetti, Tommaso Ranzani, Giada Gerboni, Thrishantha Nanayakkara,
  Kaspar Althoefer, Prokar Dasgupta, and Arianna Menciassi.
\newblock Soft robotics technologies to address shortcomings in today's
  minimally invasive surgery: the {STIFF-FLOP} approach.
\newblock {\em Soft Robotics}, 1(2):122--131, 2014.

\bibitem{clarke2018learning}
Samuel Clarke, Travers Rhodes, Christopher~G Atkeson, and Oliver Kroemer.
\newblock Learning audio feedback for estimating amount and flow of granular
  material.
\newblock In {\em Conference on Robot Learning}, pages 529--550, 2018.

\bibitem{coetzee2017calibration}
CJ~Coetzee.
\newblock Calibration of the discrete element method.
\newblock {\em Powder Technology}, 310:104--142, 2017.

\bibitem{cundall1979discrete}
Peter~A Cundall and Otto~DL Strack.
\newblock A discrete numerical model for granular assemblies.
\newblock {\em geotechnique}, 29(1):47--65, 1979.

\bibitem{dierichs2013robotic}
Karola Dierichs, Tobias Schwinn, and Achim Menges.
\newblock Robotic pouring of aggregate structures.
\newblock In {\em Rob| Arch 2012}, pages 196--205. Springer, 2013.

\bibitem{do2017automated}
Huy~Q Do, Alejandro~M Arag{\'o}n, and Dingena~L Schott.
\newblock Automated discrete element method calibration using genetic and
  optimization algorithms.
\newblock In {\em EPJ Web of Conferences}, volume 140, page 15011. EDP
  Sciences, 2017.

\bibitem{dunatunga2015continuum}
Sachith Dunatunga and Ken Kamrin.
\newblock Continuum modelling and simulation of granular flows through their
  many phases.
\newblock {\em Journal of Fluid Mechanics}, 779:483--513, 2015.

\bibitem{hu2018moving}
Yuanming Hu, Yu~Fang, Ziheng Ge, Ziyin Qu, Yixin Zhu, Andre Pradhana, and
  Chenfanfu Jiang.
\newblock A moving least squares material point method with displacement
  discontinuity and two-way rigid body coupling.
\newblock {\em ACM Transactions on Graphics (TOG)}, 37(4):150, 2018.

\bibitem{jaeger1992physics}
Heinrich~M Jaeger and Sidney~R Nagel.
\newblock Physics of the granular state.
\newblock {\em Science}, 255(5051):1523--1531, 1992.

\bibitem{jop2006constitutive}
Pierre Jop, Yo{\"e}l Forterre, and Olivier Pouliquen.
\newblock A constitutive law for dense granular flows.
\newblock {\em Nature}, 441(7094):727, 2006.

\bibitem{kamrin2018quantitative}
Ken Kamrin.
\newblock Quantitative rheological model for granular materials: The importance
  of particle size.
\newblock {\em Handbook of Materials Modeling: Applications: Current and
  Emerging Materials}, pages 1--24, 2018.

\bibitem{kamrin2019non}
Ken Kamrin.
\newblock Non-locality in granular flow: Phenomenology and modeling approaches.
\newblock {\em Frontiers in Physics}, 7:116, 2019.

\bibitem{kamrin2014effect}
Ken Kamrin and Georg Koval.
\newblock Effect of particle surface friction on nonlocal constitutive behavior
  of flowing granular media.
\newblock {\em Computational Particle Mechanics}, 1(2):169--176, 2014.

\bibitem{knuth2012discrete}
Margaret~A Knuth, JB~Johnson, MA~Hopkins, RJ~Sullivan, and JM~Moore.
\newblock Discrete element modeling of a {M}ars {E}xploration {R}over wheel in
  granular material.
\newblock {\em Journal of Terramechanics}, 49(1):27--36, 2012.

\bibitem{li2013terradynamics}
Chen Li, Tingnan Zhang, and Daniel~I Goldman.
\newblock A terradynamics of legged locomotion on granular media.
\newblock {\em science}, 339(6126):1408--1412, 2013.

\bibitem{liang2018gpu}
Jacky Liang, Viktor Makoviychuk, Ankur Handa, Nuttapong Chentanez, Miles
  Macklin, and Dieter Fox.
\newblock {GPU}-accelerated robotic simulation for distributed reinforcement
  learning.
\newblock {\em arXiv preprint arXiv:1810.05762}, 2018.

\bibitem{liu2010granular}
Andrea~J Liu and Sidney~R Nagel.
\newblock Granular and jammed materials.
\newblock {\em Soft Matter}, 6(13):2869--2870, 2010.

\bibitem{macklin2019non}
Miles Macklin, Kenny Erleben, Matthias M{\"u}ller, Nuttapong Chentanez, Stefan
  Jeschke, and Viktor Makoviychuk.
\newblock Non-smooth {N}ewton methods for deformable multi-body dynamics.
\newblock {\em ACM Transactions on Graphics (TOG)}, 38(5):1--20, 2019.

\bibitem{macklin2013position}
Miles Macklin and Matthias M{\"u}ller.
\newblock Position based fluids.
\newblock {\em ACM Transactions on Graphics (TOG)}, 32(4):104, 2013.

\bibitem{macklin2014unified}
Miles Macklin, Matthias M{\"u}ller, Nuttapong Chentanez, and Tae-Yong Kim.
\newblock Unified particle physics for real-time applications.
\newblock {\em ACM Transactions on Graphics (TOG)}, 33(4):153, 2014.

\bibitem{maladen2011undulatory}
Ryan~D Maladen, Yang Ding, Paul~B Umbanhowar, and Daniel~I Goldman.
\newblock Undulatory swimming in sand: experimental and simulation studies of a
  robotic sandfish.
\newblock {\em The International Journal of Robotics Research}, 30(7):793--805,
  2011.

\bibitem{mazhar2014studying}
Hammad Mazhar, Jonas Bollmann, Endrina Forti, Andreas Praeger, Tim Osswald, and
  Dan Negrut.
\newblock Studying the effect of powder geometry on the selective laser
  sintering process.
\newblock {\em Society of Plastics Engineers (SPE) ANTEC}, 2014.

\bibitem{muller2007position}
Matthias M{\"u}ller, Bruno Heidelberger, Marcus Hennix, and John Ratcliff.
\newblock Position based dynamics.
\newblock {\em Journal of Visual Communication and Image Representation},
  18(2):109--118, 2007.

\bibitem{negrut2013investigating}
Dan Negrut, Daniel Melanz, Hammad Mazhar, David Lamb, Paramsothy Jayakumar, and
  Michael Letherwood.
\newblock Investigating through simulation the mobility of light tracked
  vehicles operating on discrete granular terrain.
\newblock {\em SAE International Journal of Passenger Cars-Mechanical Systems},
  6(2013-01-1191):369--381, 2013.

\bibitem{nemat2002constitutive}
Sia Nemat-Nasser and Juhua Zhang.
\newblock Constitutive relations for cohesionless frictional granular
  materials.
\newblock {\em International Journal of Plasticity}, 18(4):531--547, 2002.

\bibitem{papamakarios2016fast}
George Papamakarios and Iain Murray.
\newblock Fast $\varepsilon$-free inference of simulation models with
  {B}ayesian conditional density estimation.
\newblock In {\em Advances in Neural Information Processing Systems}, pages
  1028--1036, 2016.

\bibitem{podlozhnyuk2017efficient}
Alexander Podlozhnyuk, Stefan Pirker, and Christoph Kloss.
\newblock Efficient implementation of superquadric particles in discrete
  element method within an open-source framework.
\newblock {\em Computational Particle Mechanics}, 4(1):101--118, 2017.

\bibitem{rahimi2008random}
Ali Rahimi and Benjamin Recht.
\newblock Random features for large-scale kernel machines.
\newblock In {\em Advances in Neural Information Processing Systems}, pages
  1177--1184, 2008.

\bibitem{ramos2019bayessim}
Fabio Ramos, Rafael~Carvalhaes Possas, and Dieter Fox.
\newblock Bayes{S}im: adaptive domain randomization via probabilistic inference
  for robotics simulators.
\newblock {\em arXiv preprint arXiv:1906.01728}, 2019.

\bibitem{richard2005slow}
Patrick Richard, Mario Nicodemi, Renaud Delannay, Philippe Ribiere, and Daniel
  Bideau.
\newblock Slow relaxation and compaction of granular systems.
\newblock {\em Nature materials}, 4(2):121, 2005.

\bibitem{schenck2017learning}
Connor Schenck, Jonathan Tompson, Dieter Fox, and Sergey Levine.
\newblock Learning robotic manipulation of granular media.
\newblock {\em arXiv preprint arXiv:1709.02833}, 2017.

\bibitem{sinapov2014learning}
Jivko Sinapov, Connor Schenck, and Alexander Stoytchev.
\newblock Learning relational object categories using behavioral exploration
  and multimodal perception.
\newblock In {\em 2014 IEEE International Conference on Robotics and Automation
  (ICRA)}, pages 5691--5698. IEEE, 2014.

\bibitem{thompson2015soft}
Elliot Thompson-Bean, Oliver Steiner, and Andrew McDaid.
\newblock A soft robotic exoskeleton utilizing granular jamming.
\newblock In {\em 2015 IEEE International Conference On Advanced Intelligent
  Mechatronics (AIM)}, pages 165--170. IEEE, 2015.

\bibitem{zhang2017microscopic}
Qiong Zhang and Ken Kamrin.
\newblock Microscopic description of the granular fluidity field in nonlocal
  flow modeling.
\newblock {\em Physical review letters}, 118(5):058001, 2017.

\bibitem{zhang2013ground}
Tingnan Zhang, Feifei Qian, Chen Li, Pierangelo Masarati, Aaron~M Hoover, Paul
  Birkmeyer, Andrew Pullin, Ronald~S Fearing, and Daniel~I Goldman.
\newblock Ground fluidization promotes rapid running of a lightweight robot.
\newblock {\em The International Journal of Robotics Research}, 32(7):859--869,
  2013.

\bibitem{zhu2019data}
Yifan Zhu, Laith Abdulmajeid, and Kris Hauser.
\newblock A data-driven approach for fast simulation of robot locomotion on
  granular media.
\newblock In {\em 2019 International Conference on Robotics and Automation
  (ICRA)}, pages 7653--7659. IEEE, 2019.

\end{thebibliography}
\end{document}